\documentclass[runningheads]{llncs}
\usepackage{graphicx}
\usepackage{amsmath}
\usepackage{amssymb}
\usepackage{subcaption}
\usepackage{hyperref}
\usepackage{todonotes}
\usepackage{multirow}
\usepackage{array}
\usepackage{makecell}
\usepackage[export]{adjustbox}
\usepackage{tabularx}
\newcolumntype{Y}{>{\centering\arraybackslash}X}
\newcolumntype{j}{X}
\newcolumntype{s}{>{\hsize=.15\hsize}Y}
\newcolumntype{?}{!{\vrule width 1pt}}

\begin{document}
\title{The Pitfalls of Sample Selection:\\A Case Study on Lung Nodule Classification}

\titlerunning{The Pitfalls of Sample Selection in Lung Nodule Classification}
%
%
\author{Vasileios Baltatzis \inst{1, 2}, Kyriaki-Margarita Bintsi \inst{2}, Lo\"ic Le Folgoc \inst{2},  \\  Octavio E. Martinez Manzanera \inst{1}, Sam Ellis \inst{1}, Arjun Nair \inst{3}, Sujal Desai \inst{4}, \\ Ben Glocker \inst{2}, Julia A. Schnabel \inst{1, 5, 6}}

\authorrunning{V. Baltatzis et al.}
%
\institute{School of Biomedical Engineering and Imaging Sciences, King’s College London, UK \\
\and BioMedIA, Department of Computing, Imperial College London, UK \\
\and Department of Radiology, University College London, UK \\
\and The Royal Brompton \& Harefield NHS Foundation Trust, London, UK \\
\and Technical University of Munich, Germany \\
\and Helmholtz Center Munich, Germany \\
\email{vasileios.baltatzis@kcl.ac.uk}\\ 
}
\maketitle              
\begin{abstract}
Using publicly available data to determine the performance of methodological contributions is important as it facilitates reproducibility and allows scrutiny of the published results. In lung nodule classification, for example, many works report results on the publicly available LIDC dataset. In theory, this should allow a direct comparison of the performance of proposed methods and assess the impact of individual contributions. When analyzing seven recent works, however, we find that each employs a different data selection process, leading to largely varying total number of samples and ratios between benign and malignant cases. As each subset will have different characteristics with varying difficulty for classification, a direct comparison between the proposed methods is thus not always possible, nor fair. We study the particular effect of truthing when aggregating labels from multiple experts. We show that specific choices can have severe impact on the data distribution where it may be possible to achieve superior performance on one sample distribution but not on another. While we show that we can further improve on the state-of-the-art on one sample selection, we also find that on a more challenging sample selection, on the same database, the more advanced models underperform with respect to very simple baseline methods, highlighting that the selected data distribution may play an even more important role than the model architecture. This raises concerns about the validity of claimed methodological contributions. We believe the community should be aware of these pitfalls and make recommendations on how these can be avoided in future work.

\end{abstract}

\section{Introduction}
Lung nodule characterization is the most difficult step in the pipeline of lung cancer diagnosis according to radiologists, which can be observed by a great inter-observer disagreement on the task \cite{Lin2017,Nair2018VariableRecommendations}. A lung nodule is normally characterized with respect to texture, spiculation, lobulation, and its morphological appearance on a CT scan, and eventually it must be classified as either benign or malignant for patient management. The Lung imaging Reporting And Data System (Lung-RADS) \cite{McKee2016} is a protocol that defines explicit guidelines for nodule management and follow up planning, and classifies pulmonary nodules in six categories, each of which has its own suggested follow up. Lung-RADS also integrates the PanCan Model \cite{McWilliams2013ProbabilityCT}, which provides a malignancy probability based on the morphology of a nodule and additional patient information.  Certain diagnosis can only be made through biopsy, which, however, is invasive and not always feasible to have access to. While determining the malignancy of a nodule from its appearance on a CT scan is not a fail-proof method, it is still a very useful step of the lung cancer detection pipeline. It can have very important value to clinicians in conjunction with patient history and demographics.

Several deep learning methods have been proposed for automated nodule classification from CT. The publicly available Lung Image Database Consortium and Image Database Resource Initiative (LIDC) database \cite{Armato2011,McNitt-Gray2007} has been in the core of the majority of such efforts. The LIDC does not primarily contain pathology confirmed ground truths (besides a very small subset of cases), but rather radiologists' annotations. Nevertheless, it is still heavily used by the research community for the task of lung nodule classification. Interestingly, there are various design choices regarding sample selection that need to be considered, which can have severe impact on the reported results.

The contributions of this paper can be summarized as follows: 1) We analyze several published works reporting results on LIDC nodule classification and examine the different assumptions such as annotation aggregations methods, removal of cases based on clinical guidelines, and data augmentation, which all can affect the resulting sample selection process; 2) Through an extensive experimental analysis, we show that the selected data distribution can affect the difficulty of the task and may play an even more important role than the model architecture; 3) We demonstrate that reproducibility and direct model comparison is virtually impossible to achieve and provide suggestions towards making this feasible in future work, while also making our data selection publicly available to promote reproducibility. We illustrate the pitfalls of sample selection with a novel methodological approach of curriculum by smoothing for lung nodule classification. Our findings and insights will be of use to the community and aid in the design of future approaches for lung nodule classification.

\section{State-of-the-art in Lung Nodule Classification}

\begin{figure*}[t]
\centering
\captionsetup[subfigure]{labelformat=empty}
  \begin{subfigure}[t]{0.20\linewidth}
    \centering
    \includegraphics[width =\textwidth]{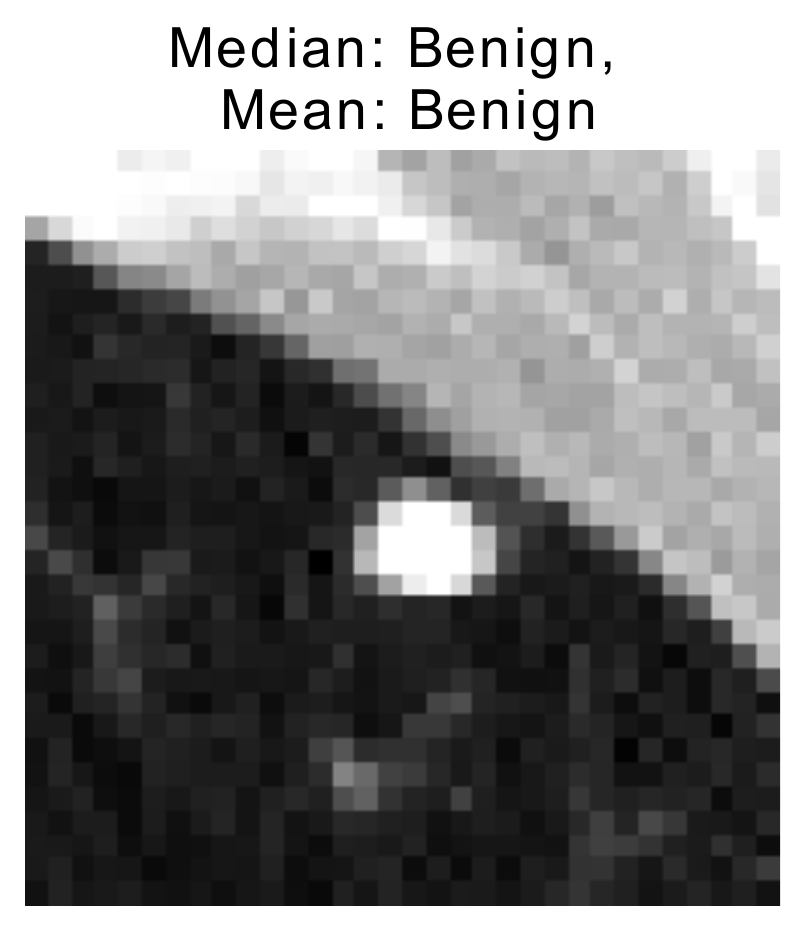}\hfill%
    \caption{} 
    \label{nod:a} 
  \end{subfigure} 
  \begin{subfigure}[t]{0.20\linewidth}
    \centering
    \includegraphics[width =\textwidth]{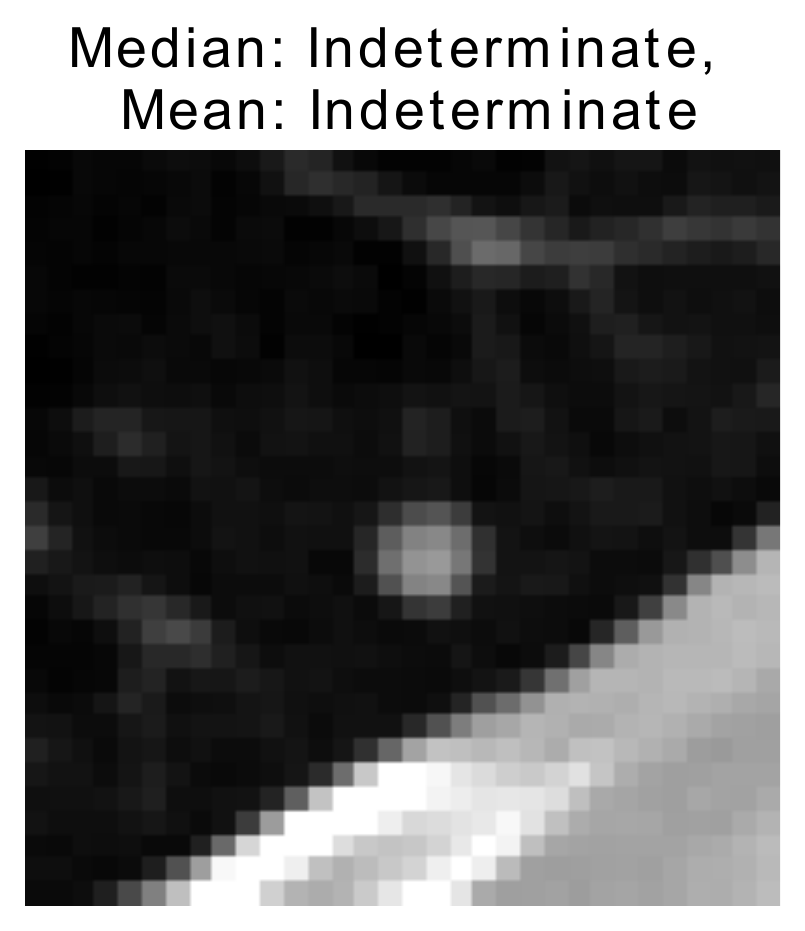}
    \caption{} 
    \label{nod:b} 
  \end{subfigure}
  \begin{subfigure}[t]{0.20\linewidth}
    \centering
    \includegraphics[width =\textwidth]{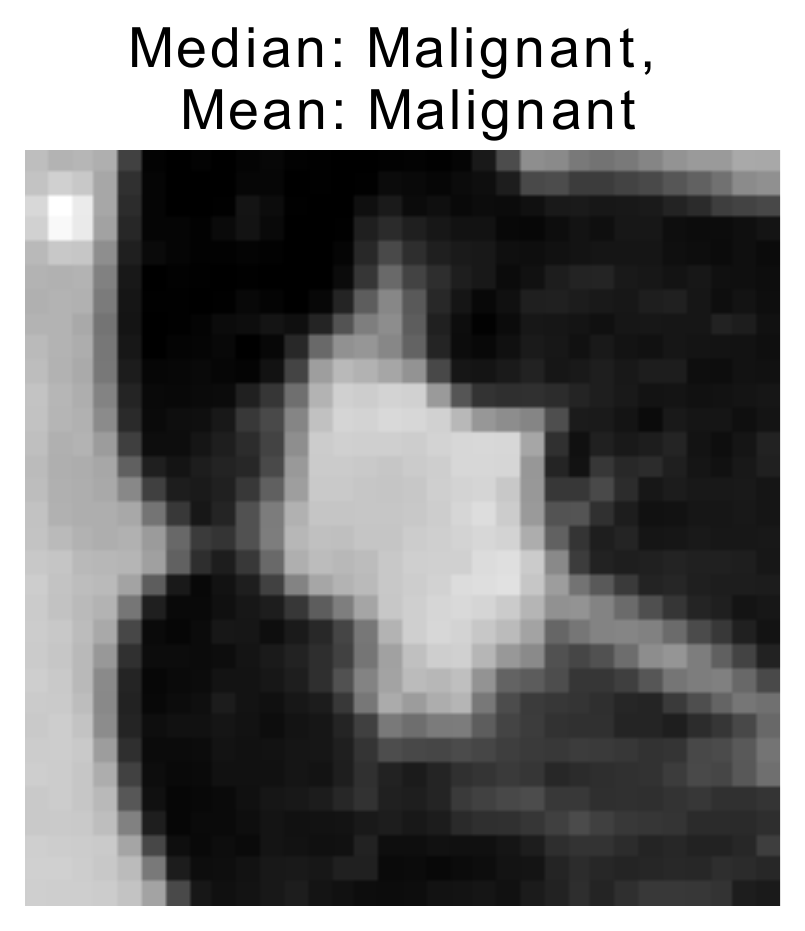}
    \caption{} 
    \label{nod:c} 
  \end{subfigure}\\
  \begin{subfigure}[t]{0.20\linewidth}
    \centering
    \includegraphics[width =\textwidth]{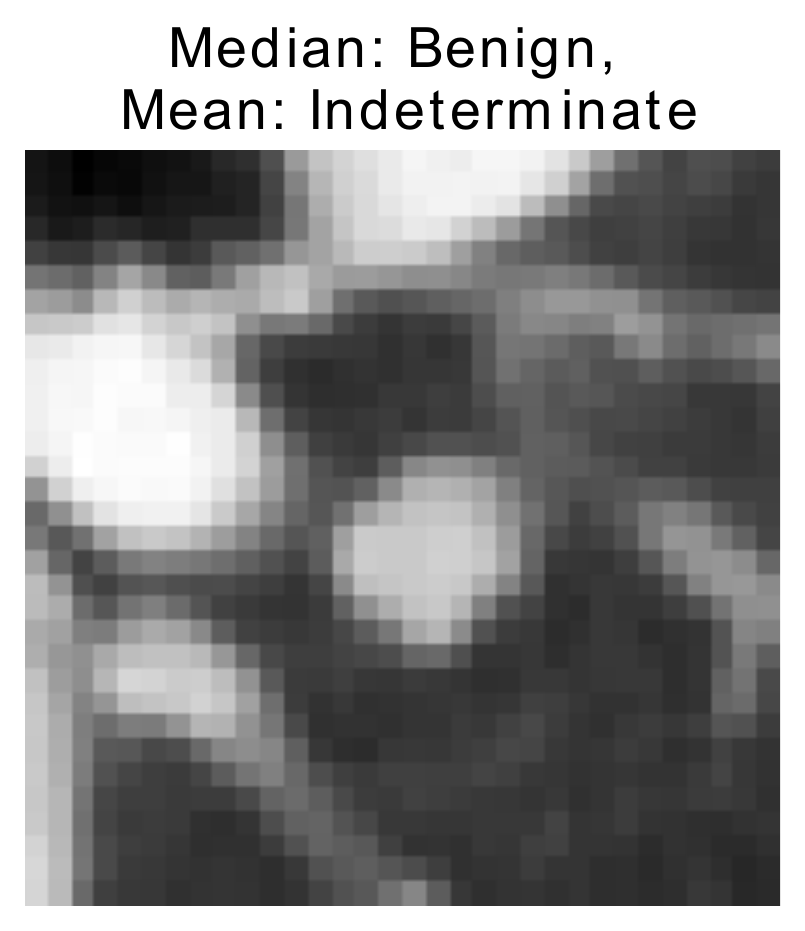}
    \caption{} 
    \label{nod:d} 
  \end{subfigure} 
    \begin{subfigure}[t]{0.20\linewidth}
    \centering
    \includegraphics[width =\textwidth]{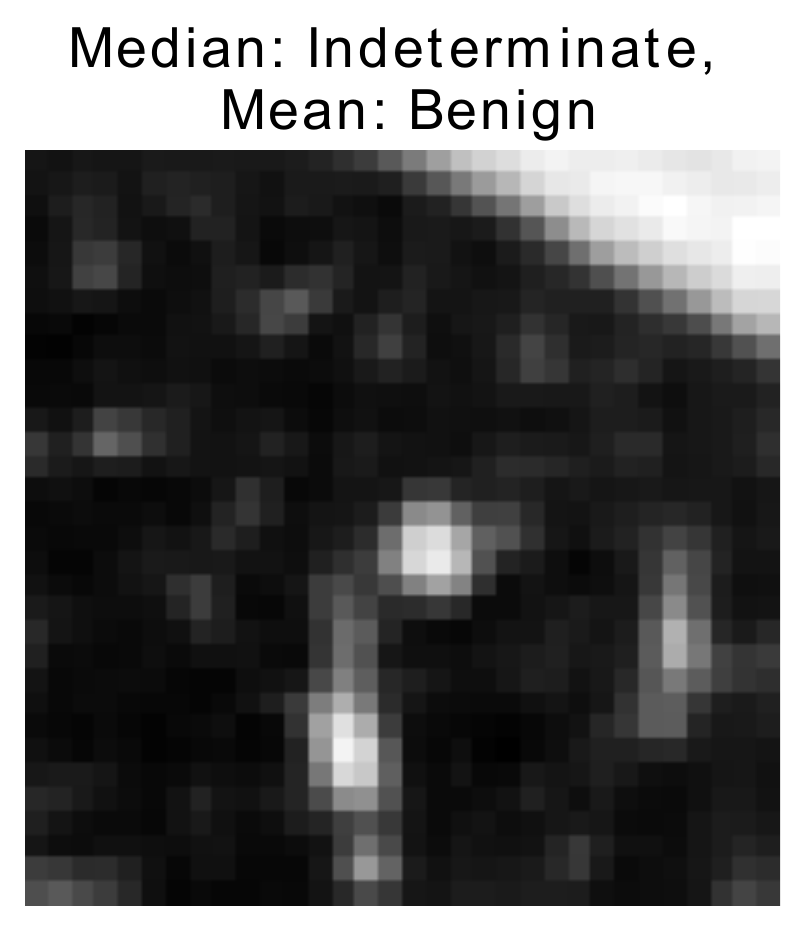}
    \caption{} 
    \label{nod:e} 
  \end{subfigure} 
    \begin{subfigure}[t]{0.20\linewidth}
    \centering
    \includegraphics[width =\textwidth]{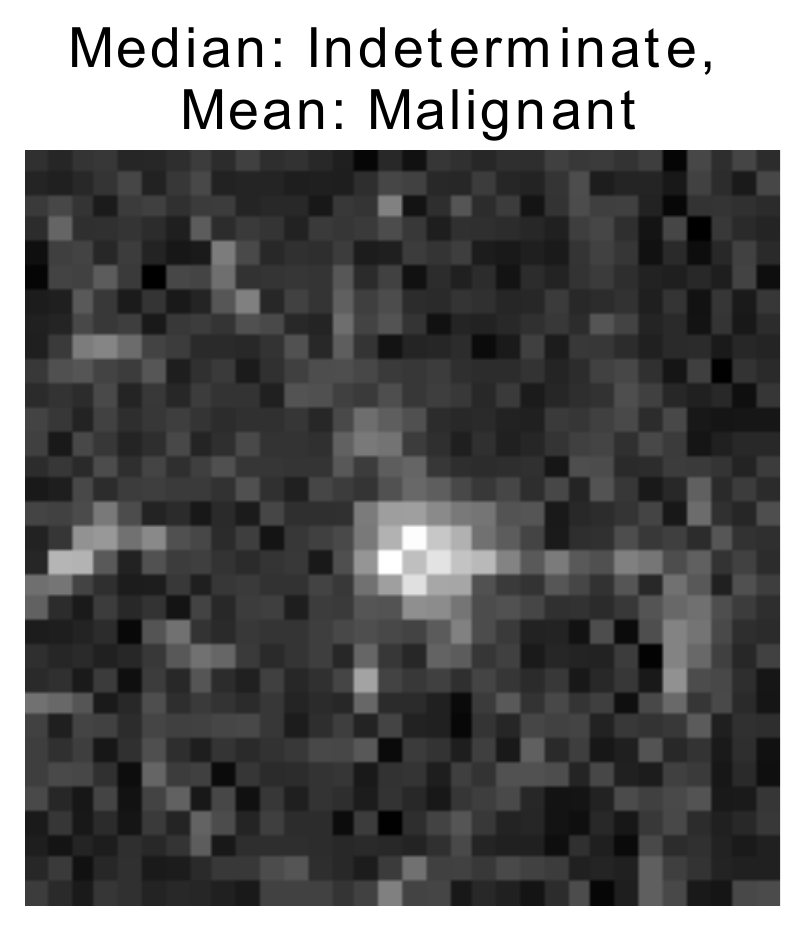}
    \caption{} 
    \label{nod:f} 
  \end{subfigure}
  \begin{subfigure}[t]{0.20\linewidth}
    \centering
    \includegraphics[width =\textwidth]{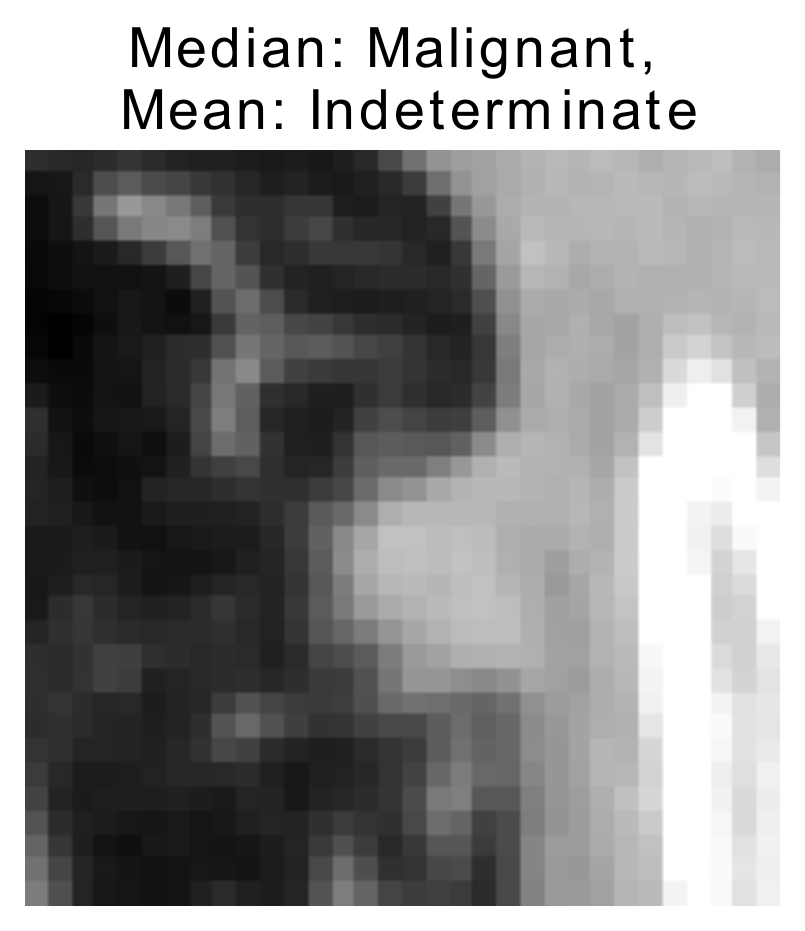}
    \caption{} 
    \label{nod:g} 
  \end{subfigure}
  \caption{Lung nodule examples from the LIDC. Top row: Nodules that have the same consensus regardless of the aggregation method used. Bottom row: Nodules that have different consensus depending on the aggregation method.}
  \label{nodules} 
  
\end{figure*}

The LIDC dataset contains more than 1000 scans. Each scan was reviewed by four radiologists who pinpointed lesion locations and assigned a variety of annotations including malignancy. For every nodule, each radiologist had to assign a malignancy rating from 1 (most likely benign) to 5 (most likely malignant). Nodules annotated with 3 were regarded as \textit{\textit{indeterminate}}.

There are a number of preprocessing and data curation steps which are considered fixed when using the LIDC and almost all recent deep learning papers follow them. These include (1) retaining only nodules that have been annotated by at least three radiologists and (2) discarding nodules annotated as \textit{indeterminate}. Subsequently, for each nodule a consensus annotation is extracted from the individual annotations through some form of aggregation or truthing (typically using mean, median, or majority voting). Example nodules from the LIDC with different consensus/aggregation combinations can be seen in Figure~\ref{nodules}. Given these relatively straightforward steps, it may be surprising to find that every paper we studied reports largely varying numbers for benign and malignant nodules and overall cases (see Table~\ref{tab1}). Most studies report that they follow a procedure similar to previous work, however, rarely provide the exact details about either the sample selection process or the final dataset (e.g. by publishing a list of scan series IDs). Beside the differences in absolute numbers of benign and malignant cases, the characteristics of the underlying data distribution may change significantly. One of the most important characteristics is the size of a nodule (quantified by its diameter), as it plays an essential role in malignancy classification. Another discrepancy arises from the decision to remove cases that have a slice thickness $>2.5 mm$, which is based on clinical guidelines \cite{Kazerooni2014ACR-STR4}. Images with thick slices are deemed unsuitable for lung cancer screening. This step was first suggested in the LUNA16 nodule detection challenge \cite{Setio2017} and has also been adopted by other studies \cite{Zhu2018}. One of the few works that release their pre-processed data is by Al-Shabi et al. \cite{Al-Shabi2019LungNetworks}. 

\begin{figure*}[t]
\centering
\begin{adjustbox}{minipage=\linewidth,scale=1}
  \begin{subfigure}[t]{0.49\linewidth}
    \centering
    \includegraphics[width =\textwidth]{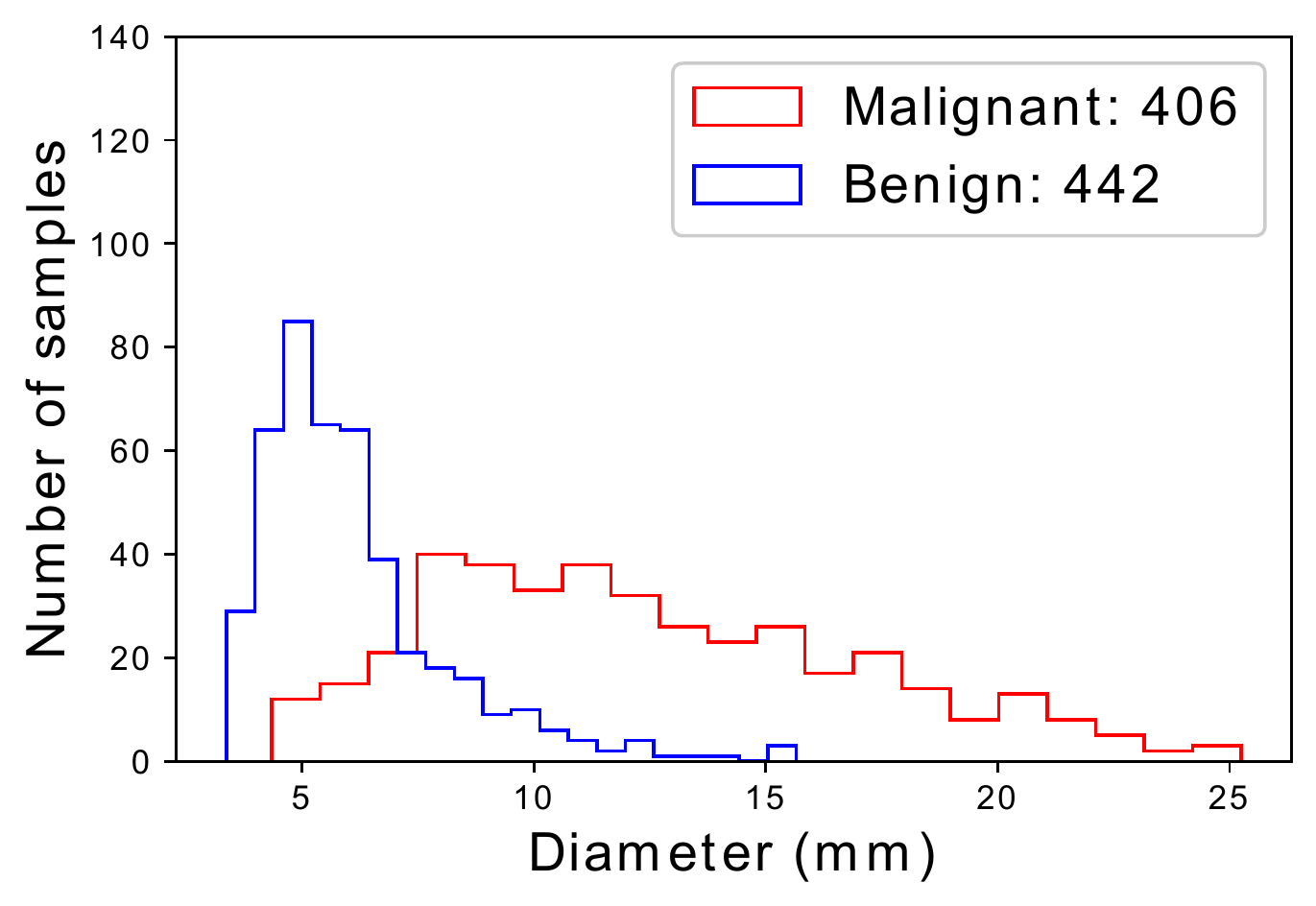}\hfill%
    \caption{} 
    \label{hist:a} 
  \end{subfigure}
  \begin{subfigure}[t]{0.49\linewidth}
    \centering
    \includegraphics[width =\textwidth]{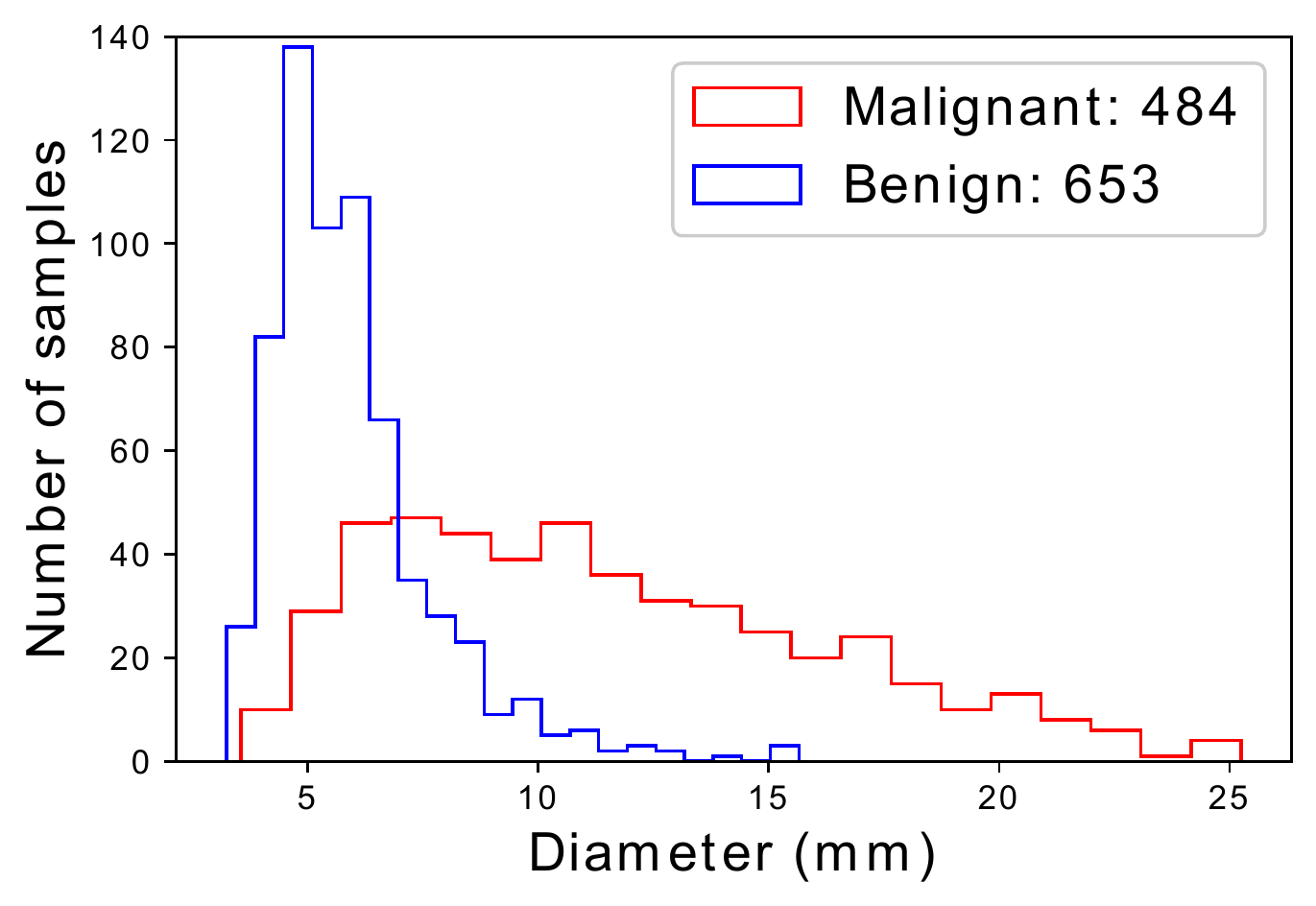}
    \caption{} 
    \label{hist:b} 
  \end{subfigure} \\
  \begin{subfigure}[t]{0.49\linewidth}
    \centering
    \includegraphics[width =\textwidth]{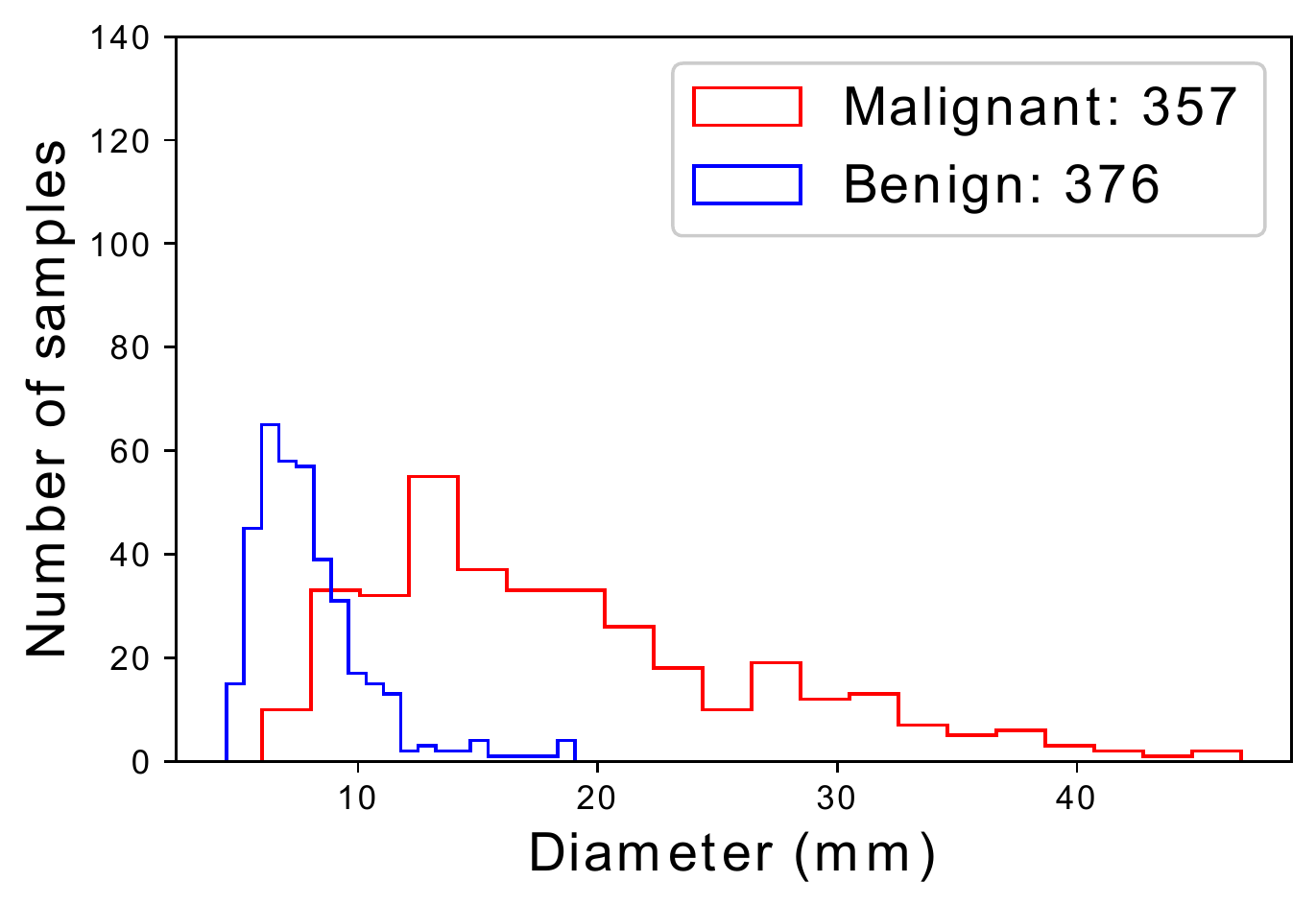}\hfill%
    \caption{} 
    \label{hist:c} 
  \end{subfigure}
  \begin{subfigure}[t]{0.49\linewidth}
    \centering
    \includegraphics[width =\textwidth]{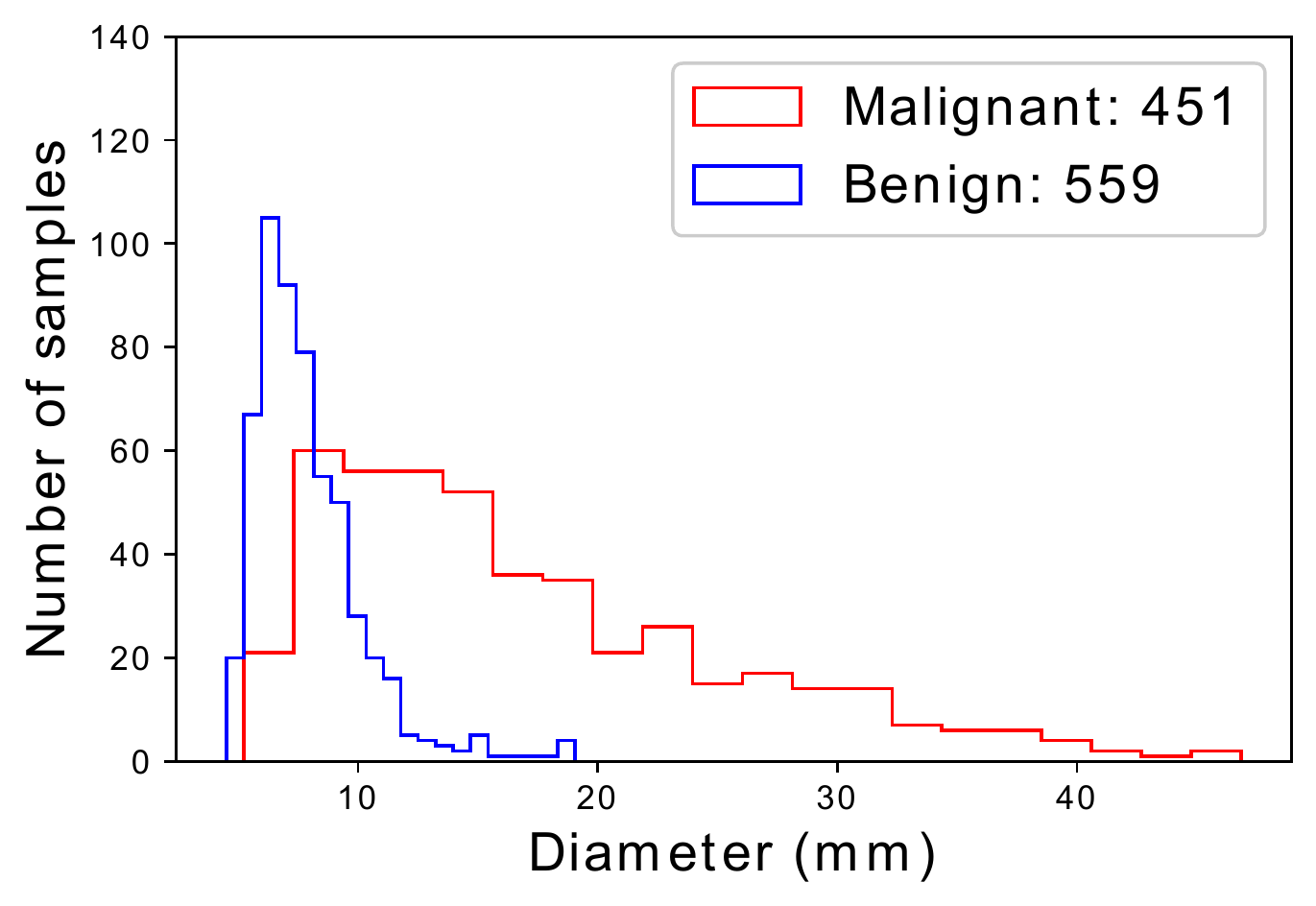}
    \caption{} 
    \label{hist:d} 
  \end{subfigure} 
    \end{adjustbox}
  \caption{Data distributions of benign and malignant samples over nodule diameter. (\subref{hist:a}) Median aggregation from \cite{Al-Shabi2019LungNetworks}, (\subref{hist:b}) Mean aggregation from \cite{Al-Shabi2019LungNetworks}, (\subref{hist:c}) Our median aggregation, (\subref{hist:d}) Our mean aggregation. Median aggregation produces fewer nodules in total (i.e. more nodules are classified as \textit{indeterminate}) for both cases, and at the same time more balanced datasets.}
  \label{hists} 

\end{figure*}

Here, we attempt to draw a direct comparison to their work with the dataset we have extracted from pre-processing LIDC (see Figure~\ref{hists}). Something like this is not feasible for the other proposed methods which do not publicly release their sample selection. In this comparison, we want to highlight the important role that the aggregation method (mean vs median) plays in determining which samples are labeled as benign and malignant. When median aggregation is used, we see that a lot more nodules have an \textit{indeterminate} consensus (i.e. median=3) and are therefore excluded, resulting in a smaller, more balanced dataset, which is much easier to separate based on the key characteristic of nodule diameter. Specifically, median aggregation leads to 442/406 benign/malignant nodules for \cite{Al-Shabi2019LungNetworks} and 376/357 benign/malignant in our replicated pipeline, respectively. In contrast, mean aggregation results in 653/484 benign/malignant for \cite{Al-Shabi2019LungNetworks} and 559/451 for us. A factor leading to a discrepancy between the two samples, even when the same aggregation method is used, is the fact that cases with a slice thickness $>2.5mm$ have been retained by \cite{Al-Shabi2019LungNetworks}. These factors make reproducibility and direct comparison of methods nearly impossible. 

\begin{table}[t]
\centering
\caption{Overview of previous work for lung nodule classification on LIDC-IDRI in terms of nodule counts and performance. Despite all papers using the same publicly available dataset, final numbers of benign and malignant cases vary largely making a direct comparison of the methods' performance impossible.}\label{tab1}
\begin{tabular}{|l|c|c|c|}
\hline
 Method &  Benign count & Malignant count & Accuracy (\%)\\
\hline
Local-Global \cite{Al-Shabi2019LungNetworks} &  442 & 406 & 89.75\\
DeepLung \cite{Zhu2018} &  554 & 450 & 90.44\\
Lightweight multi-CNN \cite{Sahu2019AEstimation} & 857 & 448 & 93.18\\
Interpretable hierarchical CNN \cite{Shen2019} & 3212 & 1040 & 84.20\\
NoduleX \cite{Causey2018} & 394 & 270 & 93.20\\
Multi-crop CNN \cite{Shen2017a} & 880 & 495 & 87.14\\
Multi-task w/ margin ranking loss \cite{Liu2020Multi-TaskAnalysis}& 972 & 450 & 93.50 \\
\hline
\end{tabular}
\end{table}

\section{Methodology}

Here we present different methods and approaches, including our attempted contribution, which we considered for studying the impact of sample selection on lung nodule classification performance. We used several baselines and state-of-the-art deep learning approaches.

\subsection{Diameter-based baselines}
\subsubsection{Diameter threshold}
The first baseline we set is not learning-based but a rather simplistic one. Specifically, given that the size of a nodule is a primary factor in determining whether a nodule is malignant or not (i.e. large nodules are most likely to be regarded by experts as malignant, while small nodules as benign) we use the provided diameter annotation in LIDC and specify a threshold for classifying nodules into benign and malignant. This baseline is used as a surrogate to determine the difficulty of the classification, as the overall size difference between structures may be easily picked up by an image-based prediction model such as a convolutional neural network (CNN).

\subsubsection{Regressed diameter threshold}
Another baseline that we use is similar to the previous one but with a CNN that is trained to regress the diameter through a mean squared error loss. The classification is taking place by applying the threshold determined from the first baseline on the output of the CNN instead of the annotation. Again, if this baseline works well, one may conclude that the task given a specific dataset is not very difficult.

\subsection{ShallowNet}
We also implement a CNN for malignancy classification (termed ShallowNet), which is a bare-bones CNN comprising of four convolutional layers with kernels of shape 3x3 and ReLU activations, and corresponding max-pooling layers with kernels of shape 2x2, as well as a fully-connected layer with 1024 neurons at the end for the classification. This is a deliberately simplistic deep learning baseline used to compare with more complicated architectures proposed in the literature.

\subsection{Local-Global}
Since we have access to the sample selection of \cite{Al-Shabi2019LungNetworks}, it makes sense to use the state-of-the-art method on this distribution. The Local-Global network was proposed by \cite{Al-Shabi2019LungNetworks} and consists of two blocks. Each block contains the following sequence: a residual sub-block \cite{He2016} followed by a non-local sub-block \cite{Wang2018Non-localNetworks} and a dropout layer. After the two blocks, there is an average pooling layer and a fully-connected layer for the classification.

\subsection{Curriculum by smoothing}
Finally, we propose the use of curriculum by smoothing (CBS) \cite{NEURIPS2020_f6a673f0}, which has shown promising results on computer vision classification tasks. CBS plays the role of our attempted methodological contribution on lung nodule classification. The main idea behind CBS is to apply a Gaussian smoothing kernel to the output of each convolutional layer of a CNN. We use $\theta \circledast x$ to denote the convolution of a kernel $\theta$ with an input $x$. Typically, in a CNN, a convolution operation is followed by a non-linear \textit{activation} function as described in Equation~\ref{cnn}:
\begin{equation}
z = \textit{activation}(\theta_{\omega} \circledast x)
\label{cnn}
\end{equation}

where $\theta_{\omega}$ are the trainable parameters of a convolutional layer.
The CBS formulation is presented in Equation~\ref{cbs}:  

\begin{equation}
z = \textit{activation}(\theta_{G} \circledast (\theta_{\omega} \circledast x))
\label{cbs}
\end{equation}
where $\theta_{G}$ is a predefined Gaussian kernel.
The Gaussian kernel is deterministic and is not trained. During the early stages of training it has an initial standard deviation $\sigma$, which is annealed as training progresses. This way, high-frequency information is suppressed in the early training steps of the CNN and is only considered at later stages of the training process. 

It is important to note that while we introduce CBS here as an approach that could enhance the performance of ShallowNet or Local-Global for the task of lung nodule classification, our purpose is not to propose a novel model architecture but rather to explore whether the selected sample distribution can play a more important role than the model architecture and highlight the pitfalls that occur in such a scenario.
\section{Experimental Analysis}

Following from the differences in the data distributions, we move to comparing some baseline models, as well as the proposed method from \cite{Al-Shabi2019LungNetworks}. In this section we focus on two distributions to demonstrate the impact of sample selection and understand whether performance differences stem from the data or the methods. Specifically, we use the data produced with median aggregation (Figure~\ref{hist:a}) from \cite{Al-Shabi2019LungNetworks} (henceforth denoted as $\mathcal{D}_{1}$), as this is the one the authors report results for, and mean aggregation (Figure~\ref{hist:d}) for our data (denoted as $\mathcal{D}_{2}$). We do not consider mean aggregation to be superior to median, but instead we want to study the differences in performance that are caused by this specific choice of truthing. Median aggregation leads to the two classes being more easily separated based on nodule diameter (Figures~\ref{hist:a},\ref{hist:c}), even though 5-10 mm is considered the most difficult area to separate malignant from benign nodules. In both $\mathcal{D}_{1}$ and $\mathcal{D}_{2}$, a nodule is considered benign when the consensus has a value lower than 3 and malignant when it has a value greater than 3.

CT scans with a slice thickness greater than 2.5mm are removed according to clinical guidelines \cite{Kazerooni2014ACR-STR4} and every remaining scan is resampled to 1mm isotropic resolution across all three dimensions and one 32x32 mm patch is extracted along each orthogonal plane at each nodule location. The final classification result for each nodule occurs from the averaging of the individual classification of each of its three planes. Some experiments include offline data augmentation (i.e. the size of the dataset itself is increased six-fold through the addition of nodule augmentations); these augmentations are the ones suggested by \cite{Al-Shabi2019LungNetworks} and include rotations, horizontal flips and Gaussian smoothing. For the proposed methodological contribution of employing CBS we choose 3x3 sized kernels, with an initial standard deviation $\sigma=1$ of the Gaussian smoothing kernel and an annealing of 0.5 every 5 epochs based on guidelines provided by the authors of \cite{NEURIPS2020_f6a673f0} and our own validation performance. All models are evaluated using 10-fold cross validation and the reported results are the average of the performance across the 10 folds. The networks are trained using the Adam optimizer \cite{Kingma2015Adam:Optimization} with learning rate $10^{-3}$ and binary cross-entropy loss for 50 epochs and a batch size of 256 samples. We also deploy early stopping to avoid overfitting. All experiments were conducted using PyTorch \cite{Paszke2017AutomaticPytorch}.

The results of the comparison can be found on Table~\ref{tab2}. First, we show that even separating the samples based on nodule diameter (i.e. thresholding) can achieve a quite high accuracy (85.02\% for $\mathcal{D}_{1}$ and 83.46\% for $\mathcal{D}_{2}$). In each case, we select the threshold that maximizes training accuracy. The threshold for the two cases is quite different (7.2mm for $\mathcal{D}_{1}$ and 11.5mm for $\mathcal{D}_{2}$) because of the different aggregation methods used and also because the equivalent diameter (i. e. the diameter of the sphere having the same volume as the nodule estimated volume) is the one used in \cite{Al-Shabi2019LungNetworks}. Then we use a shallow CNN (ShallowNet) to regress the nodule diameter and use a threshold (7.7mm for $\mathcal{D}_{1}$ and 11mm for $\mathcal{D}_{2}$) on that, in order to classify the nodule. If we focus on $\mathcal{D}_{1}$, we see that a ShallowNet trained directly on malignancy can initially just outperform the diameter-based baselines (85.74\%) but its performance gets better progressively when we use either CBS (86.80\%) or offline augmentations (89.74\%) and reaches up to 90.91\% if we use both. We observe the same pattern for Local-Global \cite{Al-Shabi2019LungNetworks} which starts from 89.15\% when we do not use CBS or augmentations and eventually reaches 90.91\% when we use both. The progressive gains from CBS and augmentations that are present in $\mathcal{D}_{1}$, however, are not replicated on $\mathcal{D}_{2}$. All the methods in that case perform very similar to the diameter-based baselines with the ShallowNet being the only one that surpasses them marginally in terms of accuracy (84.35\% with augmentations).

\begin{table}[t]
\centering
\caption{Comparison of methods on the different data distribution settings. The reported results are averaged across the 10 folds. $\mathcal{D}_{1}$ is the data distribution used in \cite{Al-Shabi2019LungNetworks}, which has occurred from median aggregation, while $\mathcal{D}_{2}$ has been extracted from the LIDC by us using mean aggregation. We use accuracy (Acc), sensitivity (Sens) and specificity (Spec) to report the performance of each method and all the reported values are percentages (\%). Even from the baselines, it is evident that $\mathcal{D}_{1}$ is an easier task to solve than $\mathcal{D}_{2}$. All methods perform better when augmented with CBS for $\mathcal{D}_{1}$. In $\mathcal{D}_{2}$ all configurations perform similarly to the diameter baseline, and there is no improvement from progressively increasing the complexity of the model by adding augmentations and/or CBS.}
\label{tab2}
\begin{tabularx}{\textwidth}{|j|s|s|s||s|s|s|}
\hline
\multirow{2}{*}{Method} &  \multicolumn{3}{c||}{$\mathcal{D}_{1}$} & \multicolumn{3}{c|}{$\mathcal{D}_{2}$}\\
\cline{2-7}
{}   & Acc & Sens & Spec & Acc & Sens & Spec \\
\hline
Diameter threshold   & 85.02 & 90.14 & 80.31 & 83.46 & 69.62 & \textbf{94.63}\\
CNN-regressed diameter threshold    & 84.43   & 84.23  & 84.61  & 81.58 & 68.95 & 91.77\\
\hline
ShallowNet  & 85.74 & 77.09 & 93.67 & 83.86 & 74.94 & 91.05\\
ShallowNet + CBS &  86.80  & 78.57 & \textbf{94.35}  &82.77& 71.17 &92.12\\
ShallowNet (w/ aug) &  89.74  & 85.96 & 93.21  & \textbf{84.35} & 77.38 & 89.98\\
ShallowNet (w/ aug) + CBS & \textbf{90.91}  & 89.40 & 92.30  &82.37 & 73.61 & 89.44\\
\hline
Local-Global \cite{Al-Shabi2019LungNetworks}  &  89.15  & 89.16  & 89.14  & 82.97 & 74.72 & 89.62\\
Local-Global + CBS & 89.26 & \textbf{91.40} & 86.94 & 81.98 & 75.38 & 87.29\\
Local-Global (w/ aug) \cite{Al-Shabi2019LungNetworks} &  89.75  & 90.17 & 88.17 & 82.57 & \textbf{79.15} & 85.33\\
Local-Global (w/ aug) + CBS &  \textbf{90.91}  & 90.64 & 91.17  & 81.88 & 70.06 & 91.41\\

\hline
\end{tabularx}
\end{table}

\section{Discussion}

The LIDC dataset has been instrumental for the majority of recent works on lung nodule classification. Here, we take a critical look at the aspect of sample selection after discovering inconsistencies in the reported literature. We aimed to examine different factors that affect the performance of a model and thus the apparent value of its methodological contribution. Starting from the pre-processing steps that various studies have applied on the LIDC dataset, we observe that a number of different assumptions during the sample selection process can lead to very different resulting data distributions (Table~\ref{tab1}). Such factors are the choice of the aggregation method (e.g. median or mean), in order to extract a consensus from the multiple annotations per nodule, or the removal of certain cases which are considered as unsuitable for the task due to clinical guidelines. 

The aggregation method, in particular, plays a very important role. First, it is affecting the total number of nodules that are retained, since median aggregation leads to more nodules having an \textit{indeterminate} consensus and consequently being removed, compared to mean aggregation. It is fair to say that these nodules, which are retained in the dataset with mean aggregation, are harder examples, and therefore, the classification task that occurs from mean aggregation is more difficult. Second, the prevalence of the two classes in the dataset changes substantially, since median aggregation leads to a more balanced, and potentially more favorable for classification, dataset. 

It is easy to understand that these choices change the nature of the underlying data distribution and hence, of the classification task itself. The comparison of the performance of different methods applied on different distributions is thus complex and makes the objective assessment of the value of methodological contributions difficult, which we also demonstrate experimentally. We initially devise several baselines. The first one is a simple thresholding based on the nodule diameter annotation. A size-relevant annotation is usually a core part of a lung nodule dataset, including the LIDC, and therefore this baseline can be applicable in all future studies. In the second baseline we apply a threshold on the diameter predictions that have been regressed by a neural network. This can indicate the degree of bias that a neural network has towards associating large nodules with malignancy and small ones with a benign nature. Given the very similar performance of the ShallowNet trained on malignancy prediction itself with the ShallowNet that is trained to regress the diameter, we understand that this bias is actually quite severe. It is well documented \cite{McWilliams2013ProbabilityCT} that the size of the nodule is an important factor in determining whether a nodule is benign, but from a clinical perspective there are also other indications such as texture or spiculation, which do not seem to be picked up by the neural network. The aforementioned baselines can describe the difficulty of the task, and we suggest their adaptation by the research community working on lung nodule classification. Additionally, we intend to publicly release our sample selection and we urge the research community to do the same to promote reproducibility. 

The core argument of our paper is epitomized when we compare the performance of all methods on the two distributions. Overall, we see that on $\mathcal{D}_{1}$, adding data augmentation or increasing the complexity of the model (i.e. Local-Global instead of ShallowNet) consistently leads to a distinct increase in performance. The approach of using CBS during training results in a performance increase on every single method, outperforming marginally even the state-of-the-art (Local-Global w/ augmentations) on $\mathcal{D}_{1}$. However, on $\mathcal{D}_{2}$, all methods are bounded by the diameter threshold baseline and even CBS is not having the impact it did on $\mathcal{D}_{1}$. This highlights the pitfalls of sample selection which may lead to incorrect conclusions about the methodological contributions. If we were to report only results on $\mathcal{D}_{1}$, we may have concluded that CBS is beneficial for lung nodule classification, and even outperforms previous works.

\section{Conclusion}

In this paper we have investigated the effect of sample selection in the context of lung nodule classification using deep learning. We have investigated different factors that cause the various published studies to report completely different number of nodules, and we show experimentally that these factors explicitly affect network performance. We have demonstrated that using progressively more and more complex methods systematically improves performance on the task, if and only if the assumptions regarding the data selection process allows for it. On the other hand, if the data distribution presents a more challenging classification task, as is the case when mean aggregation for the nodule annotations is used, then model complexity or data augmentation do not offer any kind of performance boost compared to even the simplest baseline.

\section{Acknowledgments}
\label{sec:acknowledgments}
This work is funded by the King’s College London \& Imperial College London EPSRC Centre for Doctoral Training in Medical Imaging (EP/L015226/1), EPSRC grant EP/023509/1, the Wellcome/EPSRC Centre for Medical Engineering (WT 203148/Z/16/Z), and the UKRI London Medical  Imaging \& Artificial Intelligence Centre for Value Based Healthcare. The Titan Xp GPU was donated by the NVIDIA Corporation. 

\bibliographystyle{splncs04}
\bibliography{references}

\end{document}